\documentclass[10pt,twocolumn,letterpaper]{article}

\usepackage{iccv}
\usepackage{times}
\usepackage{epsfig}
\usepackage{graphicx}
\usepackage{amsmath}
\usepackage{amssymb}
\usepackage{algorithm}
\usepackage[noend]{algpseudocode}
\usepackage{rotating}
\usepackage{hvfloat}
\usepackage{lscape}
\usepackage{hhline}
\usepackage{color,soul}
\usepackage{caption}
\usepackage{subcaption}

\usepackage[breaklinks=true,bookmarks=false]{hyperref}

\iccvfinalcopy 

\def\iccvPaperID{****} 

\begin{document}

\title{HD-CNN: Hierarchical Deep Convolutional Neural Network for Large Scale
Visual Recognition}

\author{Zhicheng Yan$^\dag$,
Hao Zhang$^\ddag$,
Robinson Piramuthu$^\ast$,
Vignesh Jagadeesh$^\ast$,
Dennis DeCoste$^\ast$,
Wei Di$^\ast$,
Yizhou Yu$^\star$\\
$^\dag$University of Illinois at Urbana-Champaign,
$^\ddag$Carnegie Mellon University\\
$^\ast$eBay Research Lab,
$^\star$The University of Hong Kong
}

\maketitle

\setlength{\textfloatsep}{1em}

\begin{abstract}

In image classification, visual separability between different object categories is highly uneven, and some categories are more difficult to distinguish than others. Such difficult categories demand more dedicated classifiers. However, existing deep convolutional neural networks (CNN) are trained as flat N-way classifiers, and few efforts have been made to leverage the hierarchical structure of categories. In this paper, we introduce hierarchical deep CNNs (HD-CNNs) by embedding deep CNNs into a category hierarchy. An HD-CNN separates easy classes using a coarse category classifier while distinguishing difficult classes using fine category classifiers. During HD-CNN training, component-wise pretraining is followed by global finetuning with a multinomial logistic loss regularized by a coarse category consistency term. In addition, conditional executions of fine category classifiers  and layer parameter compression make HD-CNNs scalable for large-scale visual recognition. We achieve state-of-the-art results on both CIFAR100 and large-scale ImageNet 1000-class benchmark datasets. In our experiments, we build up three different HD-CNNs and they lower the top-1 error of the standard CNNs by $2.65\%$, $3.1\%$ and $1.1\%$, respectively. 

\end{abstract}

\begin{figure*}[ht]
\centering
\begin{subfigure}{.22\textwidth}
  \centering
  \includegraphics[width=0.97\linewidth]{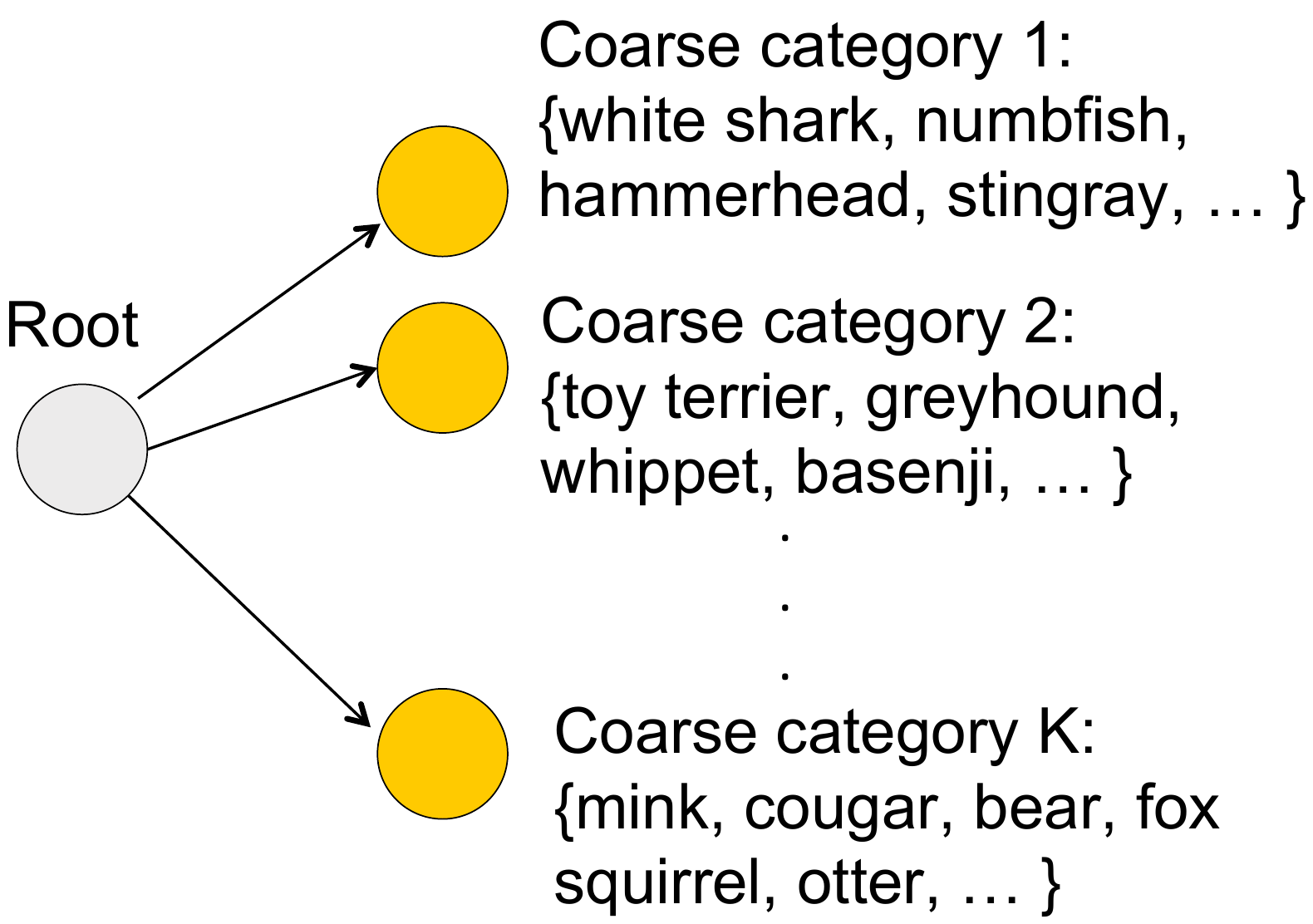}
  \caption{}
\end{subfigure}%
\begin{subfigure}{.72\textwidth}
  \centering
  \includegraphics[width=.97\linewidth]{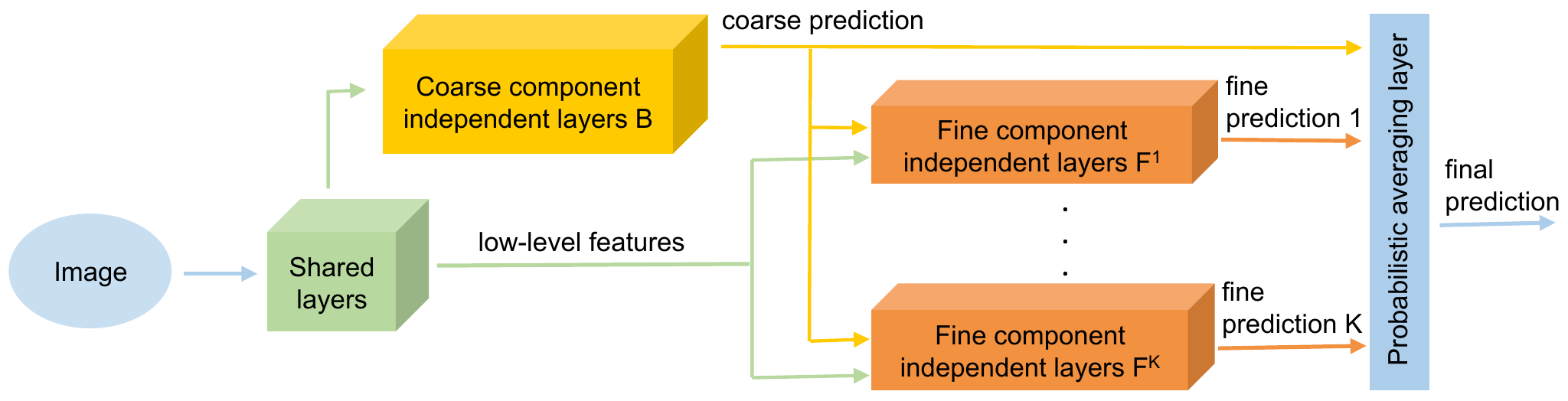}
  \caption{}
\end{subfigure}
\vspace{-2em}
\caption{(a) A two-level category hierarchy where the classes are taken from ImageNet 1000-class dataset. (b) Hierarchical Deep Convolutional Neural Network (HD-CNN) architecture.}
\label{fig:architecture}
\vspace{-2em}
\end{figure*}

\section{Introduction}


Deep CNNs are well suited for large-scale learning based visual recognition tasks because of its highly scalable training algorithm, which only needs to cache a small chunk (mini-batch) of the potentially huge volume of training data during sequential scans (epochs). They have achieved increasingly better performance in recent years. 

As datasets become bigger and the number of object categories becomes larger, one of the complications that come along is that visual separability between different object categories is highly uneven. Some categories are much harder to distinguish than others. Take the categories in CIFAR100 as an example. It is easy to tell an \textit{Apple} from a \textit{Bus}, but harder to tell an \textit{Apple} from an \textit{Orange}. In fact, both \textit{Apples} and \textit{Oranges} belong to the same coarse category \textit{fruit and vegetables} while \textit{Buses} belong to another coarse category \textit{vehicles 1}, as defined within CIFAR100.
Nonetheless, most deep CNN models nowadays are flat N-way classifiers, which share a set of fully connected layers. This makes us wonder whether such a flat structure is adequate for distinguishing all the difficult categories. A very natural and intuitive alternative organizes classifiers in a hierarchical manner according to the divide-and-conquer strategy.  Although hierarchical classification has been proven effective for conventional linear classifiers \cite{zweig2007exploiting,fergus2010semantic,zhao2011large,liu2013probabilistic}, few attempts have been made to exploit category hierarchies \cite{deng2014large,srivastava2013discriminative} in deep CNN models.

Since deep CNN models are large models themselves, organizing them hierarchically imposes the following challenges. First, instead of a handcrafted category hierarchy, how can we learn such a category hierarchy from the training data itself so that cascaded inferences in a hierarchical classifier will not degrade the overall accuracy while dedicated fine category classifiers exist for hard-to-distinguish categories? Second, a hierarchical CNN classifier consists of multiple CNN models at different levels. How can we leverage the commonalities among these models and effectively train them all? Third, it would also be slower and more memory-consuming to run a hierarchical CNN classifier on a novel testing image. How can we alleviate such limitations?

In this paper, we propose a generic and principled hierarchical architecture, \textit{Hierarchical Deep Convolutional Neural Network} (HD-CNN), that decomposes an image classification task into two steps. An HD-CNN first uses a coarse category CNN classifier to separate easy classes from one another. More challenging classes are routed downstream to fine category classifiers that focus on confusing classes. We adopt a module design principle and every HD-CNN is built upon a building block CNN. The building block can be chosen to be any of the currently top ranked single CNNs. Thus HD-CNNs can always benefit from the progress of single CNN design. An HD-CNN follows the coarse-to-fine classification paradigm and probabilistically integrates predictions from the fine category classifiers. Compared with the building block CNN, the corresponding HD-CNN can achieve lower error at the cost of a manageable increase in memory footprint and classification time.

In summary, this paper has the following contributions. First, we introduce a novel hierarchical architecture, called HD-CNN, for image classification. Second, we develop a scheme for learning the two-level organization of coarse and fine categories, and demonstrate various components of an HD-CNN can be independently pretrained. The complete HD-CNN is further fine-tuned using a multinomial logistic loss regularized by a coarse category consistency term. Third, we make the HD-CNN scalable by compressing the layer parameters and conditionally executing the fine category classifiers. We have performed evaluations on the medium-scale CIFAR100 dataset and the large-scale ImageNet 1000-class dataset, and our method has achieved state-of-the-art performance on both of them.

\section{Related Work}
\label{sec:related}
Our work is inspired by progresses in CNN design and efforts on integrating a category hierarchy with linear classifiers. The main novelty of our method is a new scalable HD-CNN architecture that integrates a category hierarchy with deep CNNs.

\subsection{Convolutional Neural Networks}
CNN-based models hold state-of-the-art performance in various computer vision tasks, including image classifcation~\cite{krizhevsky2012imagenet}, object detection~\cite{girshick2014rcnn,he2014spatial}, and image parsing~\cite{farabet2013learning}.
Recently, there has been considerable interest in enhancing CNN components, including pooling layers~\cite{zeiler2013stochastic}, activation units~\cite{goodfellow2013maxout,springenberg2013improving}, and nonlinear layers~\cite{LinCY13}. These enhancements either improve CNN training~\cite{zeiler2013stochastic}, or expand the network learning capacity. This work boosts CNN performance from an orthogonal angle and does not redesign a specific part within any existing CNN model. Instead, we design a novel generic hierarchical architecture that uses an existing CNN model as a building block. We embed multiple building blocks into a larger hierarchical deep CNN.

\subsection{Category Hierarchy for Visual Recognition}
In visual recognition, there is a vast literature exploiting category hierarchical structures \cite{tousch2012semantic}. For classification with a large number of classes using linear classifiers, a common strategy is to build a hierarchy or taxonomy of classifiers so that the number of classifiers evaluated given a testing image scales sub-linearly in the number of classes~\cite{bengio2010label,gao2011discriminative}. The hierarchy can be either predefined\cite{marszalek2007semantic,verma2012learning,jia2013visual} or learnt by top-down and bottom-up approaches~\cite{salakhutdinov2011learning, griffin2008learning,marszalek2008constructing,li2010building,bannour2012hierarchical,deng2011fast,sivic2008unsupervised}. In \cite{deng2012hedging}, the predefined category hierarchy of ImageNet dataset is utilized to achieve the trade-offs between classification accuracy and specificity. In \cite{liu2013probabilistic}, a hierarchical label tree is constructed to probabilistically combine predictions from leaf nodes. Such hierarchical classifier achieves significant speedup at the cost of certain accuracy loss.

One of the earliest attempts to introduce a category hierarchy in CNN-based methods is reported in \cite{srivastava2013discriminative} but their main goal is transferring knowledge between classes to improve the results for classes with insufficient training examples. In \cite{deng2014large}, various label relations are encoded in a hierarchy. Improved accuracy is achieved only when a subset of training images are relabeled with internal nodes in the hierarchical class tree. They are not able to improve the accuracy in the original setting where all training images are labeled with leaf nodes. In~\cite{xiao2014error}, a hierarchy of CNNs is introduced but they experimented with only two coarse categories mainly due to scalability constraints. HD-CNN exploits the category hierarchy in a novel way that we embed deep CNNs into the hierarchy in a scalable manner and achieves superior classification results over the standard CNN.

\section{Overview of HD-CNN}
\label{sec:overview}

\subsection{Notations}
The following notations are used below. A dataset consists of images $\{\mathbf{x}_i,y_i\}_{i}$. $\mathbf{x}_i$ and $y_i$ denote the image data and label, respectively. There are $C$ fine categories of images in the dataset $ \{S^f_j\}_{j=1}^C$. We will learn a category hierarchy with $K$ coarse categories $ \{S^c_{k}\}_{k=1}^{K} $.

\subsection{HD-CNN Architecture}

HD-CNN is designed to mimic the structure of category hierarchy where fine categories are divided into coarse categories as in Fig~\ref{fig:architecture}(a). It performs end-to-end classification as illustrated in Fig~\ref{fig:architecture}(b). It mainly comprises four parts, namely shared layers, a single coarse category component $B$, multiple fine category components $\{F^{k}\}_{k=1}^{K} $ and a single probabilistic averaging layer. On the left side of Fig~\ref{fig:architecture} (b) are the shared layers. They receive raw image pixel as input and extract low-level features. The configuration of shared layers are set to be the same as the preceding layers in the building block net.

On the top of Fig~\ref{fig:architecture}(b) are independent layers of coarse category component $B$ which reuses the configuration of rear layers from the building block CNN and produces a fine prediction $\{B^{f}_{ij}\}_{j=1}^{C}$ for an image $\mathbf{x_i}$. To produce a prediction $\{B_{ik}\}_{k=1}^{K}$ over coarse categories, we append a fine-to-coarse aggregation layer which aggregates fine predictions into coarse ones when a mapping from fine categories to coarse ones $P: [1,C] \mapsto [1,K]$ is given. The coarse category probabilities serve two purposes. First, they are used as weights for combining the predictions made by fine category components. Second, when thresholded, they enable conditional executions of fine category components whose corresponding coarse probabilities are sufficiently large.

In the bottom-right of Fig~\ref{fig:architecture} (b) are independent layers of a set of fine category classifiers  $\{F^{k}\}_{k=1}^{K} $, each of which makes fine category predictions. As each fine component only excels in classifying a small set of categories, they produce a fine prediction over a partial set of categories. The probabilities of other fine categories absent in the partial set are implicitly set to zero. The layer configurations are mostly copied from the building block CNN except that in the final classification layer the number of filters is set to be the size of partial set instead of the full categories.

Both coarse category component and fine category components share common layers. The reason is three-fold. First, it is shown in ~\cite{zeiler2014visualizing} that preceding layers in deep networks response to class-agnostic low-level features such as corners and edges, while rear layers extract more class-specific features such as dog face and bird's legs. Since low-level features are useful for both coarse and fine classification tasks, we allow the preceding layers to be shared by both coarse and fine components. Second, it reduces both the total floating point operations and the memory footprint of network execution. Both are of practical significance to deploy HD-CNN in real applications. Last but not the least, it can decrease the number of HD-CNN parameters which is critical to the success of HD-CNN training.


On the right side of Fig~\ref{fig:architecture} (b) is the probabilistic averaging layer which receives fine category predictions as well as coarse category prediction and produces a weighted average as the final prediction.

\vspace{-0.5em}
\begin{equation}
\label{eqn:final_pred}
p(\mathbf{x}_i)=\frac{\sum_{k=1}^{K}B_{ik}p_{k}(\mathbf{x}_i)}{\sum_{k=1}^{K}B_{ik}}    
\end{equation}
where $B_{ik}$ is the probability of coarse category $k$ for the image $\mathbf{x}_i$ predicted by the coarse category component $B$. $p_{k}(\mathbf{x}_i)$ is the fine category prediction made by the fine category component $F^{k}$.

We stress that both coarse and fine category components reuse the layer configurations from the building block CNN. This flexible modular design allows us to choose the best module CNN as the building block, depending on the task at hand.
\section{Learning a Category Hierarchy}
\label{sec:coarse_category}

Our goal of building a category hierarchy is grouping confusing fine categories into the same coarse category for which a dedicated fine category classifier will be trained. We employ a top-down approach to learn the hierarchy from the training data.

We randomly sample a held-out set of images with balanced class distribution from the training set. The rest of the training set is used to train a building block net.
We obtain a confusion matrix $\mathbf{F}$ by evaluating the net on the held-out set. 
A distance matrix $\mathbf{D}$ is derived as $\mathbf{D}=1 - \mathbf{F}$ and its diagonal entries are set to be zero. $\mathbf{D}$ is further transformed by $\mathbf{D}=0.5 * (\mathbf{D} + \mathbf{D}^T)$ to be symmetric. The entry $\mathbf{D}_{ij}$ measures how easy it is to discriminate categories $i$ and $j$.
Spectral clustering is performed on $\mathbf{D}$ to cluster fine categories into  $K$ coarse categories. 
The result is a two-level category hierarchy representing a many-to-one mapping $P^d: [1,C] \mapsto [1,K] $ from fine to coarse categories. Here, the coarse categories are disjoint.

\noindent \textbf{Overlapping Coarse Categories}
With disjoint coarse categories, the overall classification depends heavily on the coarse category classifier. If an image is routed to an incorrect fine category classifier, then the mistake can not be corrected as the probability of ground truth label is implicitly set to zero there. Removing the separability constraint between coarse categories can make the HD-CNN less dependent on the coarse category classifier. 

Therefore, we add more fine categories to the coarse categories. For a certain fine classifier $F^{k}$, we prefer to add those fine categories $\{j\}$ that are likely to be misclassfied into the coarse category $k$. Therefore, we estimate the likelihood $u^{k}(j)$ that an image in fine category $j$ is misclassified into a coarse category $k$ on the held-out set. 

\vspace{-0.5em}
\begin{equation}
\label{eqn:misclassification_prob}
u^{k}(j)=\frac{1}{\left | S^f_j \right |} \sum_{i\in S^f_j} B^d_{ik}
\end{equation}
$B^d_{ik}$ is the coarse category probability which is obtained by aggregating fine category probabilities $\{B^f_{ij}\}_j$ according to the mapping $P^d$: $B^{d}_{ik}=\sum_{j | P^d(j)=k} B^f_{ij}$. We threshold the likelihood $u^{k}(j)$  
using a parametric variable $ u_t=(\gamma K)^{-1}   $  
and add to the partial set $S^{c}_{k}$ all fine categories $\{j\}$ such that $u^{k}(j)\geq u_t$. 
Note that each branching component gives a full set prediction when $u_t = 0$ and a disjoint set prediction when $u_t = 1.0$. With overlapping coarse categories, the category hierarchy mapping $P^d$ is extended to be a many-to-many mapping $P^o$ and the coarse category predictions are updated accordingly $B^o_{ik}=\sum_{j | k \in P^o(j)} B_{ij}$. Note the sum of $\{B^o_{ik}\}_{k=1}^{K}$ exceeds $1$ and hence we perform $L_1$ normalization. The use of overlapping coarse categories was also shown to be useful for hierarchical linear classifiers~\cite{marszalek2008constructing}.

\section{HD-CNN Training}
\label{sec:training}

As we embed fine category components into HD-CNN, the number of parameters in rear layers grows linearly in the number of coarse categories. Given the same amount of training data, this increases the training complexity and the risk of over-fitting. On the other hand, the training images within the stochastic gradient descent mini-batch are probabilistically routed to different fine category components. It requires larger mini-batch to ensure parameter gradients in the fine category components are estimated by a sufficiently large number of training samples. Large training mini-batch both increases the training memory footprint and slows down the training process. Therefore, we decompose the HD-CNN training into multiple steps instead of training the complete HD-CNN from scratch as outlined in Algorithm~\ref{alg:hdcnn_training}.


\subsection{Pretraining HD-CNN}
We sequentially pretrain the coarse category component and fine category components.
\subsubsection{Initializing the Coarse Category Component}
\label{sec:pretrain_coarse}
We first pretrain a building block CNN $F^p$ using the training set. As both the preceding and rear layers in coarse category component resemble the layers in the building block CNN, we copy the weights of $F^p$ into coarse category component for initialization purpose.


\subsubsection{Pretraining the Rear Layers of Fine Category Components}
\label{sec:pretrain_fine}

Fine category components $\{F^{k}\}_k$ can be independently pretrained in parallel. Each $F^k$ should specialize in classifying fine categories within the coarse category $S^{c}_{k}$. Therefore, the pretraining of each $F^{k}$ only uses images $\{\mathbf{x_i} | i\in S^{c}_{k} \} $ from the coarse category $S^{c}_{k}$.
The shared preceding layers are already initialized and kept fixed in this stage. For each $F^{k}$, we initialize all the rear layers except the last convolutional layer by copying the learned parameters from the pretrained model $F^p$.

\begin{algorithm}[!t]
\label{alg:hdcnn_train}
\caption{HD-CNN training algorithm}\label{euclid}
\begin{algorithmic}[1]
\Procedure{HD-CNN Training}{}

\State \textbf{Step 1:} Pretrain HD-CNN
\State \hspace{\algorithmicindent}   \textbf{Step 1.1:} Initialize coarse category component
\State \hspace{\algorithmicindent}   \textbf{Step 1.2:} Pretrain fine category components
\State \textbf{Step 2:} Fine-tune the complete HD-CNN
\EndProcedure
\end{algorithmic}
\label{alg:hdcnn_training}
\end{algorithm}
\subsection{Fine-tuning HD-CNN}
\label{sec:finetune_hdcnn}
After both coarse category component and fine category components are properly pretrained, we fine-tune the complete HD-CNN. Once the category hierarchy as well as the associated mapping $P^o$ is learnt, each fine category component focuses on classifying a fixed subset of fine categories. During fine-tuning, the semantics of coarse categories predicted by the coarse category component should be kept consistent with those associated with fine category components. Thus we add a coarse category consistency term to regularize the conventional multinomial logistic loss.

\noindent \textbf{Coarse category consistency} The learnt fine-to-coarse category mapping $P:[1,C] \mapsto [1,K]$ provides a way to specify the target coarse category distribution $\{t_{k}\}$. Specifically, $t_{k}$ is set to be the fraction of all the training images within the coarse category $S^c_k$ under the assumption the distribution over coarse categories across the training dataset is close to that within a training mini-batch.

\vspace{-0.5em}
\begin{equation}
\vspace{-0.5em}
\label{eqn:target_sparsity}
t_{k}=\frac{\sum_{j| k \in P(j)} \left | S_j \right | }{\sum_{k' =1}^{K} \sum_{j| k' \in P(j)} \left | S_j \right |  } \quad  \forall k \in [1,K]
\end{equation}

The final loss function we use for fine-tuning the HD-CNN is shown below.

\vspace{-0.5em}
\begin{equation}
\vspace{-0.5em}
\label{eqn:loss_function}
E=-\frac{1}{n}\sum_{i=1}^{n}log(p_{y_i})+ \frac{\lambda}{2}\sum_{k=1}^{K}(t_{k}-\frac{1}{n}\sum_{i=1}^{n}B_{ik})^2
\end{equation}
where $n$ is the size of training mini-batch. $\lambda$ is a regularization constant and is set to $\lambda=20$.  


\section{HD-CNN Testing}
\label{sec:hdcnn_testing}
As we add fine category components into the HD-CNN, the number of parameters, memory footprint and  execution time in rear layers,  all scale linearly in the number of coarse categories. 
To ensure HD-CNN is scalable for large-scale visual recognition, we develop conditional execution and layer parameter compression techniques.

\noindent \textbf{Conditional Execution}. At test time, for a given image, it is not necessary to evaluate all fine category classifiers as most of them have insignificant weights $B_{ik}$ as in Eqn~\ref{eqn:final_pred}. Their contributions to the final prediction are negligible. Conditional executions of the top weighted fine components can accelerate the HD-CNN classification. Therefore, we threshold $B_{ik}$ using a parametric variable $B_t=(\beta K)^{-1}$ and reset $B_{ik}$ to zero when $B_{ik}<B_t$. Those fine category classifiers with $B_{ik}=0$ are not evaluated.

\noindent \textbf{Parameter Compression}. In HD-CNN, the number of parameters in rear layers of fine category classifiers grows linearly in the number of coarse categories. Thus we compress the layer parameters at test time to reduce memory footprint. Specifically, we choose the Product Quantization approach~\cite{jegou2011product} to compress the parameter matrix $W \in R^{m\times n}$ by first partitioning it horizontally into segments of width $s$. $W=[W^1, ..., W^{(n/s)}]$. Then $k$-means clustering is used to cluster the rows in $W^i, \forall i\in [1,(n/s)]$. By only storing the nearest cluster indices in a 8-bit integer matrix $I \in R^{m\times (n/s)}$ and cluster centers in a single-precision floating number matrix $C \in R^{k\times n}$, we can achieve a compression factor $(32mn)/(32kn+8mn/s)$. The hyperparameters for parameter compression are $(s,k)$.

\section{Experiments}
\label{sec:experiments}

\subsection{Overview}
We evaluate HD-CNN on the benchmark datasets  CIFAR100~\cite{krizhevsky2009learning} and ImageNet~\cite{deng2009imagenet}. HD-CNN is implemented on the widely deployed \textit{Caffe}~\cite{Jia13caffe} software. The network is trained by back propagation \cite{krizhevsky2012imagenet}. We run all the testing experiments on a single NVIDIA Tesla K40c card.

\subsection{CIFAR100 Dataset}
\label{sec:cifar100_nin}
The CIFAR100 dataset consists of 100 classes of natural images. There are 50K training images and 10K testing images. We follow ~\cite{goodfellow2013maxout} to preprocess the datasets (e.g. global contrast normalization and ZCA whitening). Randomly cropped and flipped image patches of size $26 \times 26$ are used for training. We adopt a \textit{NIN} network \footnote{\url{https://github.com/mavenlin/cuda-convnet/blob/master/NIN/cifar-100_def}} with three stacked layers~\cite{LinCY13}. 
We denote it as CIFAR100-NIN which will be the HD-CNN building block. Fine category components share preceding layers from \textit{conv1} to \textit{pool1} which accounts for $6\%$ of the total parameters and $29\%$ of the total floating point operations. The remaining layers are used as independent layers.

For building the category hierarchy, we randomly choose 10K images from the training set as  held-out set. Fine categories within the same coarse categories are visually more similar. We pretrain the rear layers of fine category components. The initial learning rate is $0.01$ and it is decreased by a factor of 10 every 6K iterations. Fine-tuning is performed for 20K iterations with large mini-batches of size 256. The initial learning rate is $0.001$ and is reduced by a factor of 10 once after 10K iterations.

For evaluation, we use 10-view testing \cite{krizhevsky2012imagenet}. We extract five $26 \times 26$ patches (the 4 corner patches and the center patch) as well as their horizontal reflections and average their predictions. The CIFAR100-NIN net obtains $35.27\%$ testing error. Our HD-CNN achieves testing error of $32.62\%$ which improves the building block net by $2.65\%$. 

\begin{figure}[h]
\vspace{-1em}
\centering
\begin{subfigure}{.30\textwidth}
  \centering
  \includegraphics[width=1.0\linewidth]{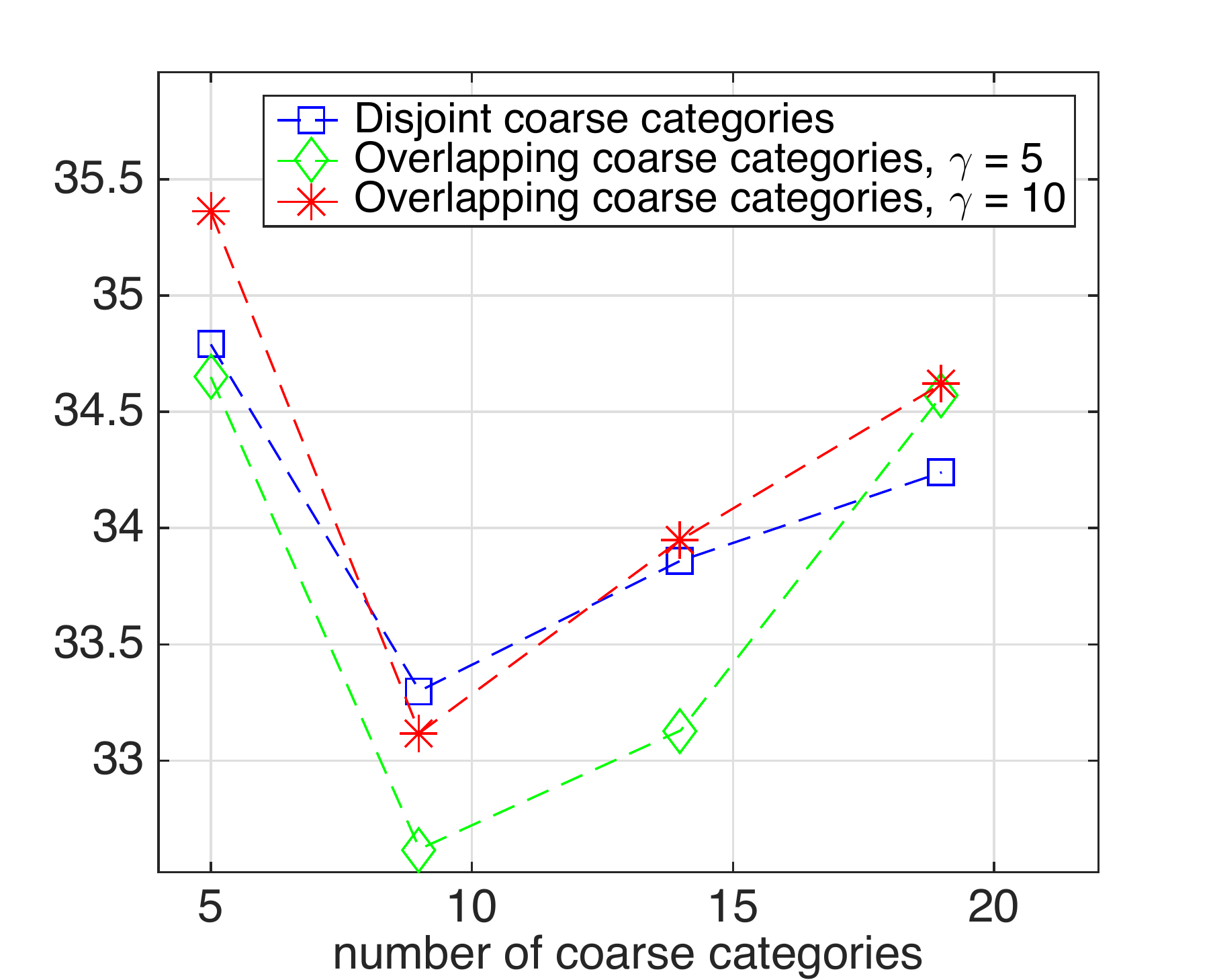}
\end{subfigure}%
\begin{subfigure}{.18\textwidth}
  \centering
  \includegraphics[width=1.0\linewidth]{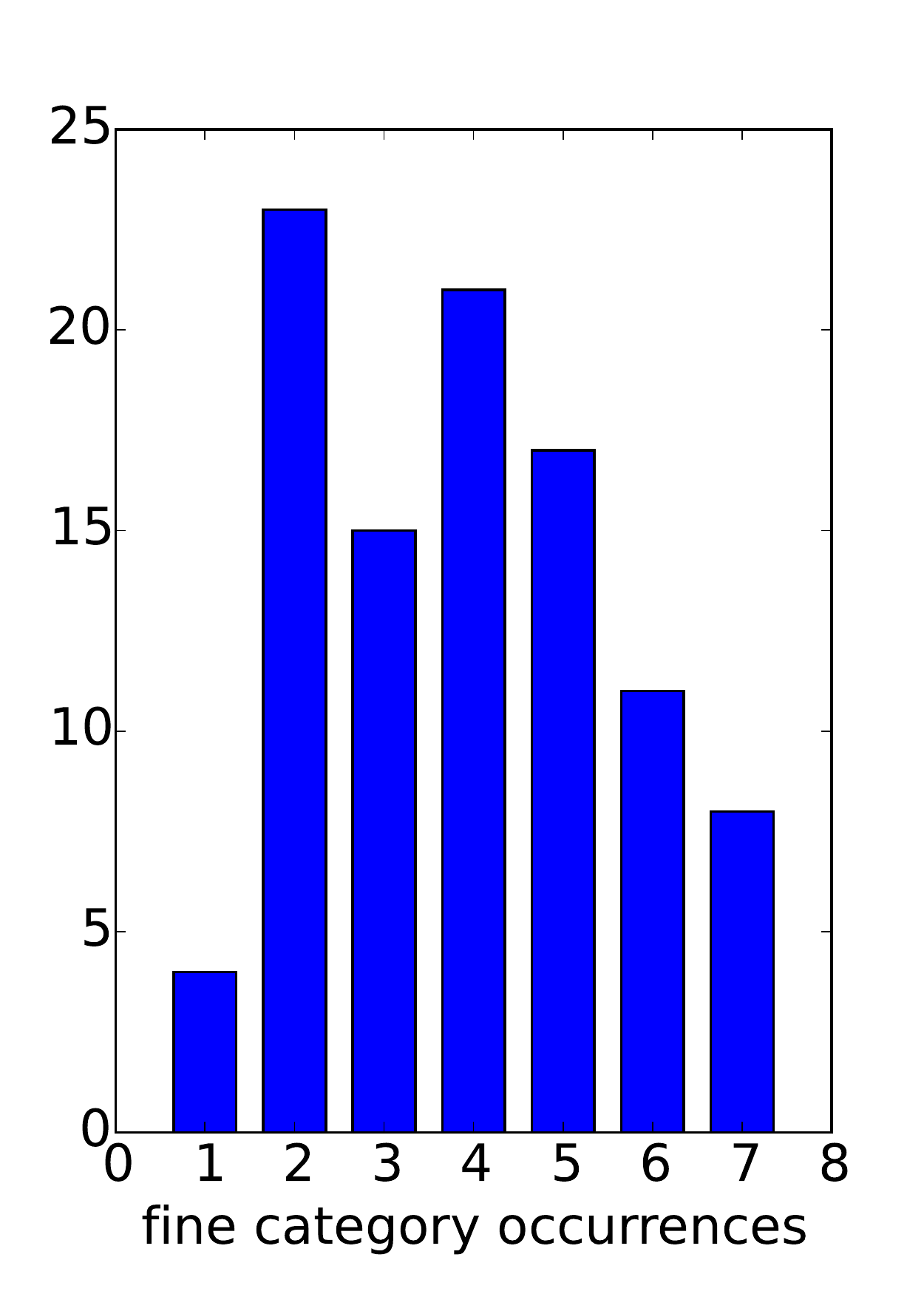}
\end{subfigure}
\caption{\textbf{Left}: HD-CNN 10-view testing error against the number of coarse categories on CIFAR100 dataset. \textbf{Right}: Histogram of fine category occurrences in 9 overlapping coarse categories.}
\vspace{-1em}
\label{fig:cifar100_internal_comp}
\end{figure}

\noindent \textbf{Category hierarchy}. 
During the construction of the category hierarchy, the number of coarse categories can be adjusted by the clustering algorithm. We can also make the coarse categories either disjoint or overlapping by varying the hyperparameter $\gamma$. Thus we investigate their impacts on the classification error. We experiment with 5, 9, 14 and 19 coarse categories and vary the value of $\gamma$. The best results are obtained with 9 overlapping coarse categories and $\gamma = 5$ as shown in Fig~\ref{fig:cifar100_internal_comp} left. A histogram of fine category occurrences in 9 overlapping coarse categories is shown in Fig \ref{fig:cifar100_internal_comp} right. 
The optimal value of coarse category number and hyperparameter $\gamma$ are dataset dependent, mainly affected by the inherent hierarchy within the categories. 

{\renewcommand{\arraystretch}{1.2}%
\begin{table}[t]
\caption{10-view testing errors on CIFAR100 dataset. Notation \textbf{CCC}=coarse category consistency. }
\label{tab: cifar_results} 
\vspace{-2em}
\begin{center}
    \begin{tabular}{ p{6.4cm} | p{1.1cm}}
    Method & Error \\ 
    \hline \hline
    Model averaging ($2$ CIFAR100-NIN nets) &  $35.13$ \\ \hline    
    DSN ~\cite{lee2014deeply}  & $34.68$  \\ \hline
    CIFAR100-NIN-double & $34.26$ \\ \hline    
    dasNet~\cite{stollenga2014deep} &  $33.78$  \\ \hline
    \hline
    Base: CIFAR100-NIN & $35.27$ \\ \hline
    HD-CNN, no finetuning &  $33.33$ \\ \hline
    HD-CNN, finetuning w/o CCC &  $33.21$ \\ \hline
    \textbf{HD-CNN, finetuning w/ CCC} &  $\mathbf{32.62}$ \\ \hline
    \end{tabular}   
\end{center}
\vspace{-1em}
\end{table}
}

\noindent \textbf{Shared layers}. The use of shared layers makes both computational complexity and memory footprint of HD-CNN sublinear in the number of fine category classifiers when compared to the building block net.  Our HD-CNN with 9 fine category classifiers based on CIFAR100-NIN consumes less than three times as much memory as the building block net without parameter compression. We also want to investigate the impact of the use of shared layers on the classification error, memory footprint and the net execution time (Table~\ref{tab: internal_comp}). We build  another HD-CNN where coarse category component and all fine category components use independent preceding layers initialized from a pretrained building block net. Under the single-view testing where only a central cropping is used, we observe a minor error increase from $34.36\%$ to $34.50\%$. But using shared layers dramatically reduces the memory footprint from $1356$ MB to $459$ MB and testing time from $2.43$ seconds to $0.28$ seconds.

\noindent \textbf{Conditional executions}. By varying the hyperparameter $\beta$, we can effectively affect the number of fine category components that will be executed. There is a trade-off between execution time and classification error. A larger value of $\beta$ leads to higher accuracy at the cost of executing more components for fine categorization. By enabling conditional executions with hyperparameter $\beta=6$, we obtain a substantial $2.8$X speed up with merely a minor increase in error from $34.36\%$ to $34.57\%$ (Table ~\ref{tab: internal_comp}). The testing time of HD-CNN is about $2.5$ times as much as that of the building block net.

\noindent \textbf{Parameter compression}. As fine category CNNs have independent layers from \textit{conv2} to \textit{cccp6}, we compress them and reduce the memory footprint from $447$MB to $286$MB with a minor increase in error from $34.57\%$ to $34.73\%$.

\noindent \textbf{Comparison with a strong baseline.}
As our HD-CNN memory footprint is about two times as much as the building block model (Table~\ref{tab: internal_comp}), it is necessary to compare a stronger baseline of similar complexity with HD-CNN. We adapt CIFAR100-NIN and double the number of filters in all convolutional layers which accordingly increases the memory footprint by three times. We denote it as CIFAR100-NIN-double and obtain error $34.26\%$ which is $1.01\%$ lower than that of the building block net but is  $1.64\%$ higher than that of HD-CNN.

\noindent \textbf{Comparison with model averaging}. HD-CNN is fundamentally different from model averaging~\cite{krizhevsky2012imagenet}. In model averaging, all models are capable of classifying the full set of the categories and each one is trained independently. The main sources of their prediction differences are different initializations. In HD-CNN, each fine category classifier only excels at classifying a partial set of categories.
To compare HD-CNN with model averaging, we independently train two CIFAR100-NIN networks and take their averaged prediction as the final prediction. We obtain an error of $35.13\%$, which is about $2.51\%$ higher than that of HD-CNN (Table \ref{tab: cifar_results}). 
Note that HD-CNN is orthogonal to the model averaging and an ensemble of HD-CNN networks can further improve the performance.

\noindent \textbf{Coarse category consistency}. To verify the effectiveness of coarse category consistency term in our loss function (\ref{eqn:loss_function}), we fine-tune a HD-CNN using the traditional multinomial logistic loss function. The testing error is $33.21\%$, which is $0.59\%$ higher than that of a HD-CNN fine-tuned with coarse category consistency (Table \ref{tab: cifar_results}). 

\noindent \textbf{Comparison with state-of-the-art.} Our HD-CNN improves on the current two best methods ~\cite{lee2014deeply} and ~\cite{stollenga2014deep} by $2.06\%$ and $1.16\%$ respectively and sets new state-of-the-art results on CIFAR100 (Table \ref{tab: cifar_results}).

\subsection{ImageNet 1000-class Dataset}
\label{sec:imagenet_nin}

The ILSVRC-2012 ImageNet dataset consists of about $1.2$ million training images, $50,000$ validation images. To demonstrate the generality of HD-CNN, we experiment with two different building block nets. In both cases, HD-CNNs achieve significantly lower testing errors than the building block nets.

\subsubsection{Network-In-Network Building Block Net}

\begin{figure*}[ht]
\begin{center}
\includegraphics[width=0.99\textwidth]{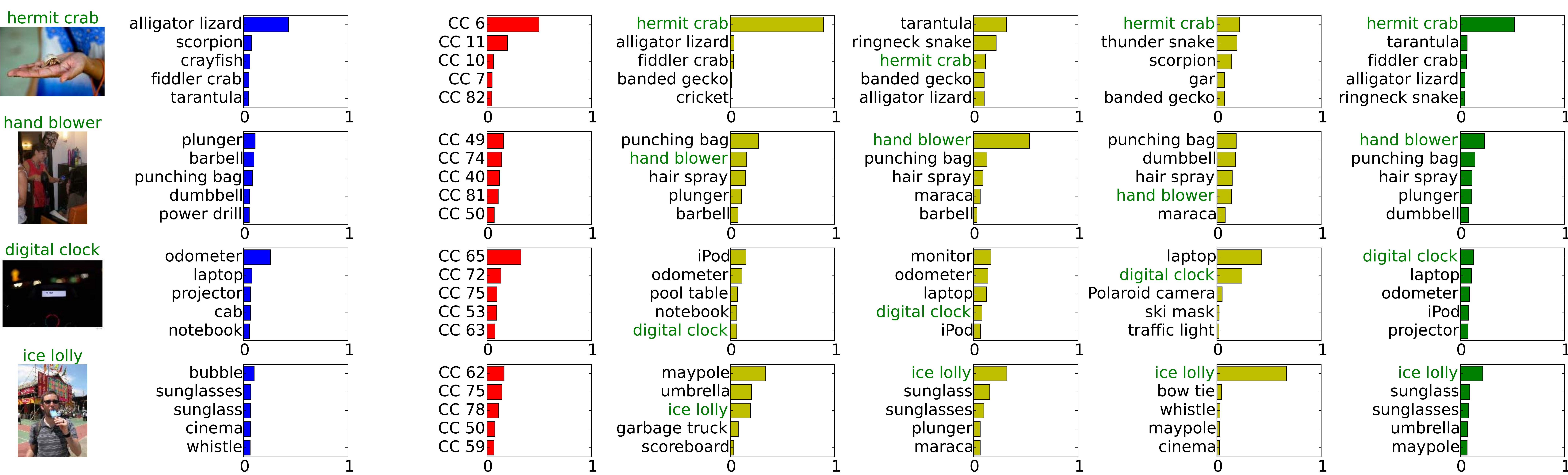}
 \centerline{(a) \hfill (b) \hfill (c) \hfill (d)\hfill (e) \hfill (f) \hfill (g)}
\end{center}
\vspace{-1em}
   \caption{Case studies on ImageNet dataset. Each row represents a testing case. \textbf{Column (a)}: test image with ground truth label. \textbf{Column (b)}: top 5 guesses from the building block net ImageNet-NIN.  \textbf{Column (c)}: top 5 Coarse Category (\textbf{CC}) probabilities. \textbf{Column (d)-(f)}: top 5 guesses made by the top 3 fine category CNN components. \textbf{Column (g)}: final top 5 guesses made by the HD-CNN.  See text for details.
}
\vspace{-2.5em}
\label{fig:case study}
\end{figure*}

We choose a public 4-layer NIN net \footnote{\url{https://gist.github.com/mavenlin/d802a5849de39225bcc6}} as our first building block as it has greatly reduced number of parameters compared to AlexNet~\cite{krizhevsky2012imagenet} but similar error rates. It is denoted as ImageNet-NIN. In HD-CNN, various components share preceding layers from \textit{conv1} to \textit{pool3} which account for $26\%$ of the total parameters and $82\%$ of the total floating point operations.

We follow the training and testing protocols as in~\cite{krizhevsky2012imagenet}. Original images are resized to $256\times 256$. Randomly cropped and horizontally reflected $224\times 224$ patches are used for training. At test time, the net makes a 10-view averaged prediction.  We train ImageNet-NIN for 45 epochs. 
The top-1 and top-5 errors are $39.76\%$ and $17.71\%$.

{\renewcommand{\arraystretch}{1.2}%
\begin{table}[ht]
\vspace{-1em}
\caption{Comparison of testing errors, memory footprint and testing time between building block nets and HD-CNNs on CIFAR100 and ImageNet datasets. Statistics are collected under \textit{single-view} testing. Three building block nets CIFAR100-NIN, ImageNet-NIN and ImageNet-VGG-16-layer are used. The testing mini-batch size is 50. Notations: \textbf{SL}=Shared layers, \textbf{CE}=Conditional executions, \textbf{PC}=Parameter compression.}
\label{tab: internal_comp} 
\vspace{-1em}
\begin{center}
    \begin{tabular}{  p{3.3cm} | p{2.0cm} | p{1.0cm} |p{0.6cm} }
    Model & top-1, top-5 & Memory (MB) & Time (sec.) \\ 
    \hline \hline
    Base:CIFAR100-NIN &  $37.29$ &  188 & 0.04 \\ \hline  
    HD-CNN w/o SL &  $34.50$ &  1356  & 2.43 \\ \hline     
    \textbf{HD-CNN} &  $\mathbf{34.36}$ &  459  & 0.28 \\ \hline     
    HD-CNN+CE &  $34.57$ &  447  & 0.10 \\ \hline 
    HD-CNN+CE+PC &  $34.73$ &  286  & 0.10 \\ \hline
    \hline
    Base:ImageNet-NIN &  $41.52,18.98$ &  535 & 0.19 \\ \hline       
	\textbf{HD-CNN} &  $\mathbf{37.92},\mathbf{16.62}$ &  3544 & 3.25 \\ \hline    
    HD-CNN+CE&  $38.16,16.75$ &  3508 & 0.52 \\ \hline    
    HD-CNN+CE+PC&  $38.39,16.89$ &  1712 & 0.53 \\ \hline
    \hline 
    Base:ImageNet-VGG-16-layer &  $32.30,12.74$ & 4134 & 1.04 \\ \hline 
    \textbf{HD-CNN+CE+PC}&  $\mathbf{31.34},\mathbf{12.26}$ &  6863 & 5.28 \\ \hline 
    \end{tabular}   
\end{center}
\vspace{-0.5em}
\end{table}
}

\begin{figure}[h]
\vspace{-0.5em}
\centering
\begin{subfigure}{.23\textwidth}
  \centering
  \includegraphics[width=1.0\linewidth]{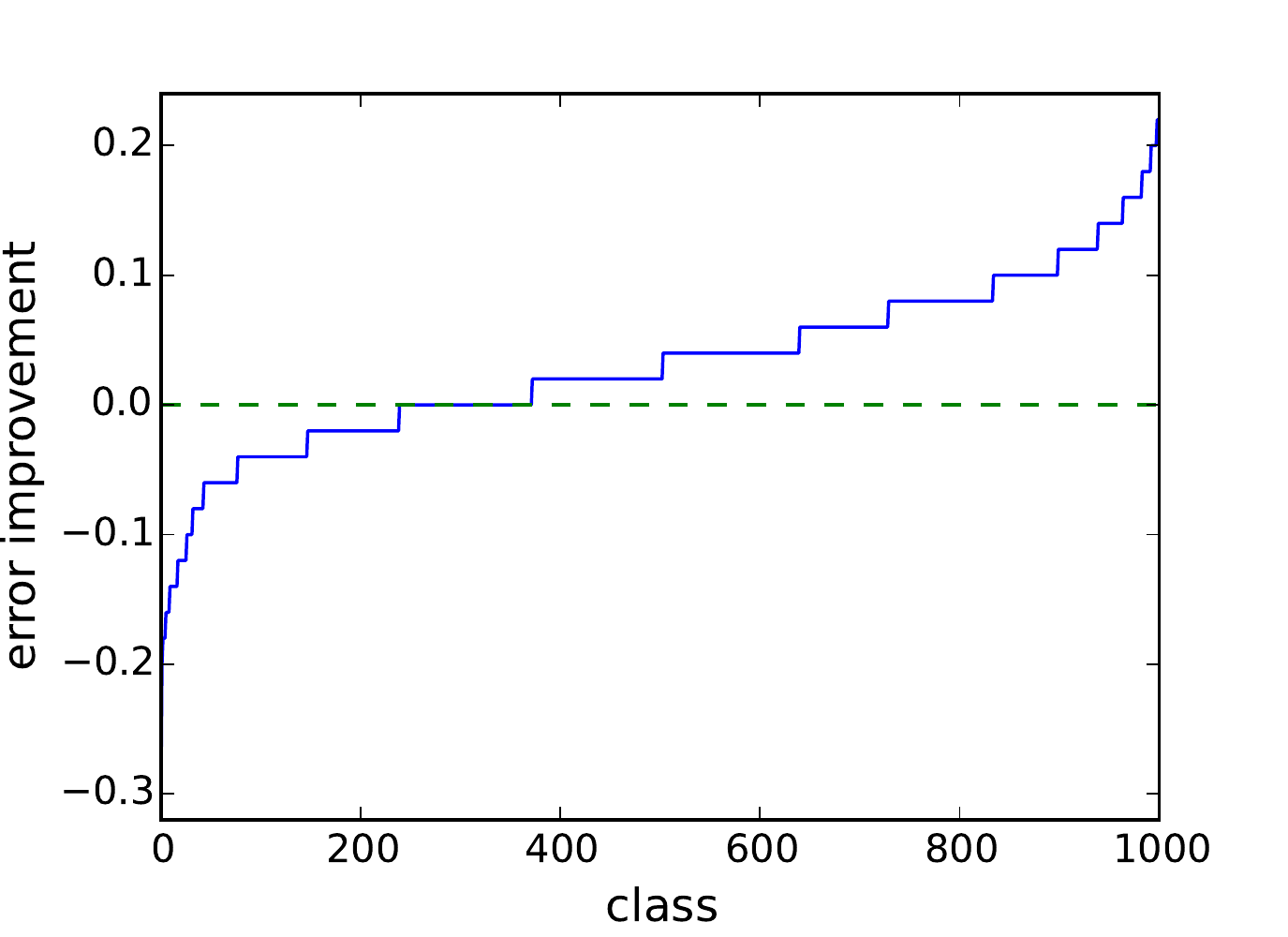}
\end{subfigure}%
\begin{subfigure}{.27\textwidth}
  \centering
  \includegraphics[width=1.0\linewidth]{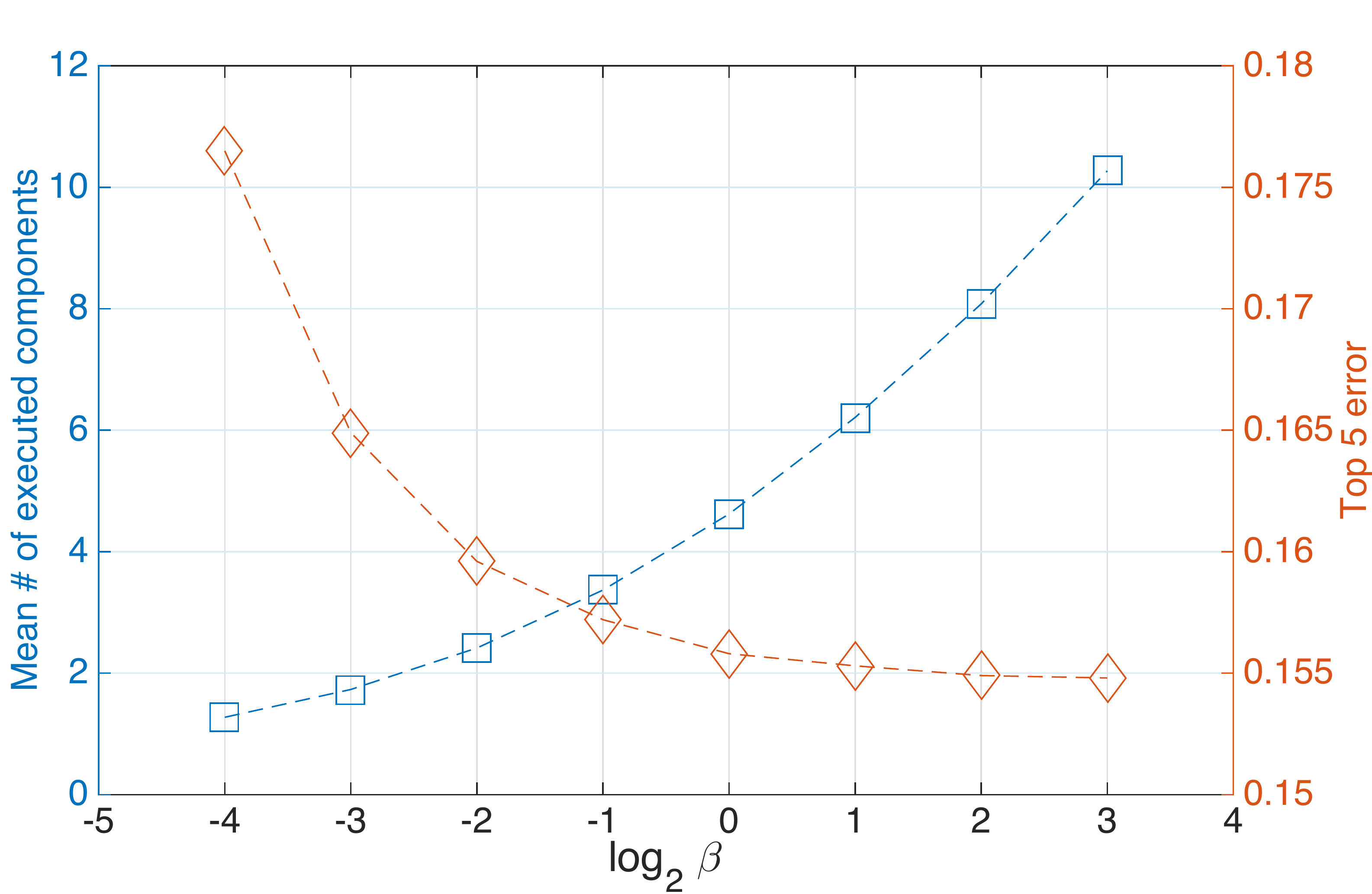}
\end{subfigure}
\caption{\textbf{Left}: Class-wise HD-CNN top-5 error improvement over the building block net. \textbf{Right}: Mean number of executed fine category classifiers and top-5 error against hyperparameter $\beta$ on the ImageNet validation dataset.}
\vspace{-1em}
\label{fig:coarse_category_nin_imagenet_conditional_exe}
\end{figure}

To build the category hierarchy, we take 100K training images as the held-out set and find 89 overlapping coarse categories. 
Each fine category CNN is fine-tuned for 40K iterations while the initial learning rate $0.01$ is decreased by a factor of $10$ every 15K iterations. Fine-tuning the complete HD-CNN is not performed as the required mini-batch size is significantly higher than that for the building block net. Nevertheless, we still achieve top-1 and top-5 errors of $36.66\%$ and $15.80\%$ and improve the building block net by $3.1\%$ and $1.91\%$, respectively (Table~\ref{tab: imagenet_accuracy_imagenet_nin}). The class-wise top-5 error improvement over the building block net is shown in Fig \ref{fig:coarse_category_nin_imagenet_conditional_exe} left.

\noindent \textbf{Case studies} We want to investigate how HD-CNN corrects the mistakes made by the building block net. In Fig \ref{fig:case study}, we collect four testing cases. In the first case, the building block net fails to predict the label of the tiny \textit{hermit crab} in the top 5 guesses. In HD-CNN, two coarse categories $\#6$ and $\#11$ receive most of the coarse probability mass. The fine category component $\#6$ specializes in classifying crab breeds and strongly suggests the ground truth label. By combining the predictions from the top fine category classifiers, the HD-CNN predicts \textit{hermit crab} as the most probable label. In the second case, the ImageNet-NIN confuses the ground truth \textit{hand blower} with other objects of close shapes and appearances, such as \textit{plunger} and \textit{barbell}. For HD-CNN, the coarse category component is also not confident about which coarse category the object belongs to and thus assigns even probability mass to the top coarse categories. For the top 3 fine category classifiers, $\#74$ strongly predicts ground truth label while the other two $\#49$ and $\#40$ rank the ground truth label at the 2nd and 4th place respectively. Overall, the HD-CNN ranks the ground truth label at the 1st place. This demonstrates HD-CNN needs to rely on multiple fine category classifiers to make correct predictions for difficult cases.


\noindent \textbf{Overlapping coarse categories}.To investigate the impact of overlapping coarse categories on the classification, we train another HD-CNN with 89 fine category classifiers using disjoint coarse categories. It achieves top-1 and top-5 errors of $38.44\%$ and $17.03\%$ respectively, which is higher than those of the HD-CNN using overlapping coarse category hierarchy by $1.78\%$ and $1.23\%$ (Table~\ref{tab: imagenet_accuracy_imagenet_nin}).

\noindent \textbf{Conditional executions}. By varying the hyperparameter $\beta$, we can control the number of fine category components that will be executed. There is a trade-off between execution time and classification error as shown in Fig  \ref{fig:coarse_category_nin_imagenet_conditional_exe} right. A larger value of $\beta$ leads to lower error at the cost of more executed fine category components. By enabling conditional executions with hyperparameter $\beta=8$, we obtain a substantial $6.3$X speed up with merely a minor increase of single-view testing top-5 error from $16.62\%$ to $16.75\%$ (Table~\ref{tab: internal_comp}). With such speedup, the HD-CNN testing time is less than $3$ times as much as that of the building block net.

\noindent \textbf{Parameter compression}. We compress independent layers \textit{conv4} and \textit{cccp7} as they account for $60\%$ of the parameters in ImageNet-NIN. Their parameter matrices are of size $1024\times 3456$ and $1024\times 1024$ and we use compression hyperparameters $(s,k)=(3,128)$ and $(s,k)=(2,256)$. The compression factors are $4.8$ and $2.7$. The compression decreases the memory footprint from $3508$MB to $1712$MB and merely increases the top-5 error from $16.75\%$ to $16.89\%$ under single-view testing (Table~\ref{tab: internal_comp}).

\noindent \textbf{Comparison with model averaging}. As the HD-CNN memory footprint is about three times as much as the building block net, we independently train three ImageNet-NIN nets and average their predictions. We obtain top-5 error $17.11\%$ which is $0.6\%$ lower than the building block but is $1.31\%$ higher than that of HD-CNN (Table~\ref{tab: imagenet_accuracy_imagenet_nin}).

{\renewcommand{\arraystretch}{1.2}%
\begin{table}[t]
\vspace{-1.0em}
\caption{Comparisons of 10-view testing errors between ImageNet-NIN and HD-CNN. Notation \textbf{CC}=Coarse category.}
\vspace{-1.5em}
\label{tab: imagenet_accuracy_imagenet_nin} 
\begin{center}
    \begin{tabular}{ p{4.3cm} |p{2.8cm}}
    Method & top-1, top-5 \\
    \hline\hline
    Base:ImageNet-NIN & $39.76$, $17.71$ \\ \hline
    Model averaging ($3$ base nets) &  $38.54$, $17.11$ \\ \hline  
	HD-CNN, disjoint CC &  $38.44, 17.03$  \\ \hline      
    \textbf{HD-CNN} &  $\mathbf{36.66}$, $\mathbf{15.80}$ \\ \hline    
    \end{tabular}   
\end{center}
\vspace{-1.5em}
\end{table}
}

\subsubsection{VGG-16-layer Building Block Net}
The second building block net we use is a 16-layer CNN from \cite{simonyan2014very}. We denote it as ImageNet-VGG-16-layer\footnote{\url{https://github.com/BVLC/caffe/wiki/Model-Zoo}}. The layers from \textit{conv1\_1} to \textit{pool4} are shared and they account for $5.6\%$ of the total parameters and $90\%$ of the total floating number operations. The remaining layers are used as independent layers in coarse and fine category classifiers. We follow the training and testing protocols as in \cite{simonyan2014very}. For training, we first sample a size $S$ from the range $[256,512]$ and resize the image so that the length of short edge is $S$. Then a randomly cropped and flipped patch of size $224\times224$ is used for training. For testing, dense evaluation is performed on three scales $\{256,384,512\}$ and the averaged prediction is used as the final prediction. Please refer to \cite{simonyan2014very} for more training and testing details. On ImageNet validation set, ImageNet-VGG-16-layer achieves top-1 and top-5 errors $24.79\%$ and $7.50\%$ respectively.

We build a category hierarchy with 84 overlapping coarse categories. We implement multi-GPU training on Caffe by exploiting data parallelism~\cite{simonyan2014very} and train the fine category classifiers on two NVIDIA Tesla K40c cards. The initial learning rate is $0.001$ and it is decreased by a factor of 10 every 4K iterations. HD-CNN fine-tuning is not performed. Due to large memory footprint of the building block net (Table~\ref{tab: internal_comp}), the HD-CNN with 84 fine category classifiers cannot fit into the memory directly. Therefore, we compress the parameters in layers \textit{fc6} and \textit{fc7} as they account for over $85\%$ of the parameters. Parameter matrices in \textit{fc6} and \textit{fc7} are of size $4096\times 25088$ and $4096\times 4096$. Their compression hyperparameters are $(s,k)=(14,64)$ and $(s,k)=(4,256)$. The compression factors are $29.9$ and $8$ respectively.
The HD-CNN obtains top-1 and top-5 errors $23.69\%$ and $6.76\%$ on ImageNet validation set and improves over ImageNet-VGG-16-layer by $1.1\%$ and $0.74\%$ respectively.

{\renewcommand{\arraystretch}{1.2}%
\begin{table}[h]
\vspace{-1.5em}
\caption{Errors on ImageNet validation set.}
\label{tab: imagenet_accuracy} 
\vspace{-1.5em}
\begin{center}
    \begin{tabular}{ p{5.7cm} |p{1.8cm}}
    Method & top-1, top-5 \\ 
    \hline \hline
    GoogLeNet,multi-crop~\cite{szegedy2014going} & N/A,$7.9$\\ \hline
    VGG-19-layer, dense~\cite{simonyan2014very} & $24.8$,$7.5$ \\ \hline 
    VGG-16-layer+VGG-19-layer,dense & $24.0$,$7.1$ \\ \hline 
    \hline
    Base:ImageNet-VGG-16-layer,dense & $24.79$,$7.50$ \\ \hline
    \textbf{HD-CNN,dense} &  $\mathbf{23.69}$,$\mathbf{6.76}$ \\ \hline   
    \end{tabular}   
\end{center}
\vspace{-1.5em}
\end{table}
}

\noindent \textbf{Comparison with state-of-the-art}. Currently, the two best nets on ImageNet dataset are GoogLeNet~\cite{szegedy2014going} (Table~\ref{tab: imagenet_accuracy}) and VGG 19-layer network~\cite{simonyan2014very}. Using multi-scale multi-crop testing, a single GoogLeNet net achieves top-5 error $7.9\%$.
With multi-scale dense evaluation, a single VGG 19-layer net obtains top-1 and top-5 errors $24.8\%$ and $7.5\%$ and improves top-5 error of GoogLeNet by $0.4\%$. Our HD-CNN decreases top-1 and top-5 errors of VGG 19-layer net by $1.11\%$ and $0.74\%$ respectively. Furthermore, HD-CNN slightly outperforms the results of averaging the predictions from
VGG-16-layer and VGG-19-layer nets.

\section{Conclusions and Future Work}
\label{ref:conclusion}
We demonstrated that HD-CNN is a flexible deep CNN architecture to improve over existing deep CNN models. We showed this empirically on both CIFAR-100 and Image-Net datasets using three different building block nets. As part of future work, we plan to extend HD-CNN architectures to those with more than 2 hierarchical levels and also verify our empirical results in a theoretical framework.



{\small
\bibliographystyle{ieee}
\bibliography{hdcnn}
}


\end{document}